%% file: main.tex
\documentclass[10pt,twocolumn,letterpaper]{article}

\usepackage[pagenumbers]{iccv} 
\usepackage{times}
\usepackage{epsfig}
\usepackage{graphicx}
\usepackage{amsmath}
\usepackage{amssymb}
\usepackage{enumitem}
\usepackage{makecell}
\usepackage{color}
\usepackage{adjustbox}
\usepackage{multirow}
\usepackage{colortbl}
\usepackage{booktabs}
\usepackage{threeparttable}

\renewcommand{\thefootnote}{\fnsymbol{footnote}}
\definecolor{iccvblue}{rgb}{0.21,0.49,0.74}
\usepackage[pagebackref,breaklinks,colorlinks,allcolors=iccvblue]{hyperref}

\def\paperID{*} 
\def\confName{ICCV}
\def\confYear{2025}
\begin{document}
\title{MambaVSR: Content-Aware Scanning State Space Model for Video Super-Resolution}


\author{
  Linfeng He\textsuperscript{\rm 1,\rm 2}\enspace 
  Meiqin Liu\textsuperscript{\rm 1,\rm 2\thanks{Corresponding author.}}\enspace 
  Qi Tang\textsuperscript{\rm 1,\rm 2}\enspace 
  Chao Yao\textsuperscript{\rm 3}\enspace 
  Yao Zhao\textsuperscript{\rm 1,\rm 2} \\
  \textsuperscript{1} Institute of Information Science, Beijing Jiaotong University \\
  \textsuperscript{2} Visual Intelligence + X International Cooperation Joint Laboratory of MOE, \\
  Beijing Jiaotong University\\ 
  \textsuperscript{3} School of Computer and Communication Engineering, \\
  University of Science and Technology Beijing\\
  \vspace{-2mm}
}

\twocolumn[{
\renewcommand\twocolumn[1][]{#1}
\maketitle
\vspace{-1.2cm}
\begin{center}
    \centering
    \includegraphics[width=1.0\linewidth]{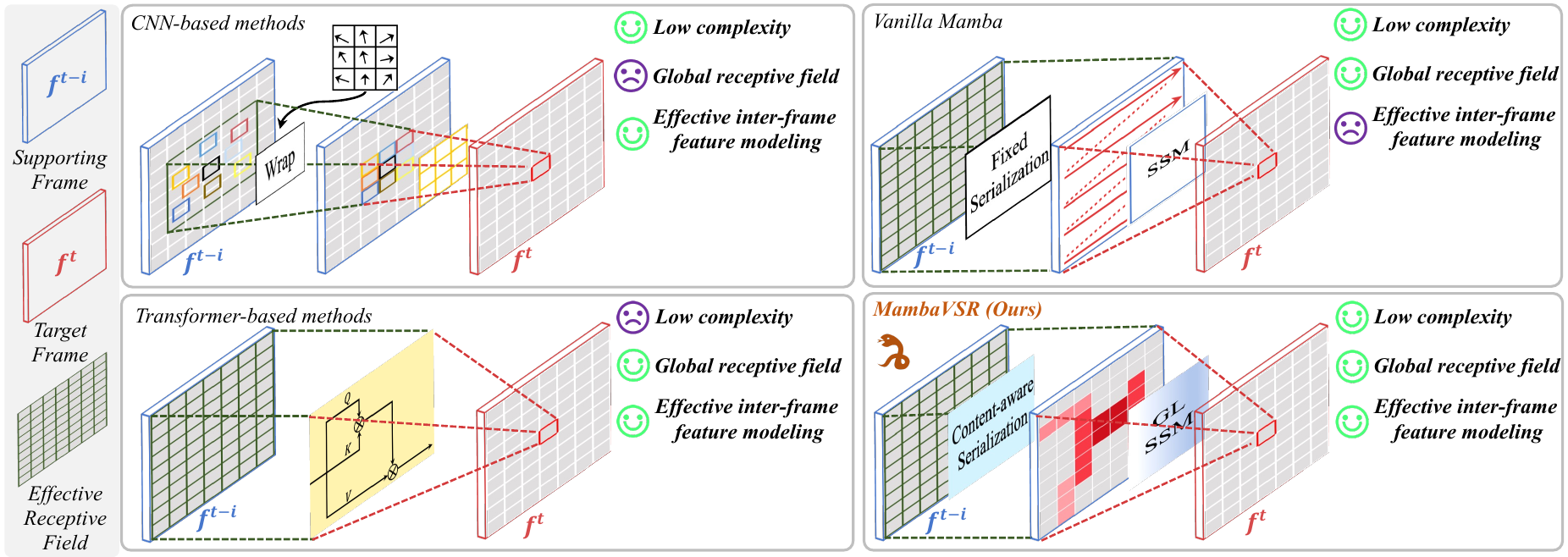}
    \captionof{figure}{Illustration of four methods in VSR tasks. {CNN-based methods} use optical flow to warp frames, capturing limited temporal context. {Transformer-based methods} exploit temporal information via global self-attention with quadratic complexity. {vanilla Mamba} fails to capture inter-frame similarities due to its inherent design. {MambaVSR} effectively models long-range dependencies and enhances local details through content-aware scanning. }
    \label{fig:intro}
\end{center}
\vspace{0.5cm}
}]

\renewcommand{\thefootnote}{\fnsymbol{footnote}}
\footnotetext[1]{corresponding author}

\maketitle
\input{sec/0_abstract}    
\input{sec/1_intro}
\input{sec/2_RelatedWork}

\input{sec/3_method}
\input{sec/4_experiments}

\input{sec/5_conclusion}
{
    \small
    \bibliographystyle{ieeenat_fullname}
    \bibliography{main}
}

\end{document}

%% file: sec/0_abstract.tex
\begin{abstract}
Video super-resolution (VSR) faces critical challenges in effectively modeling non-local dependencies across misaligned frames while preserving computational efficiency. 
Existing VSR methods typically rely on optical flow strategies or transformer architectures, which struggle with large motion displacements and long video sequences. 
To address this, we propose MambaVSR, the first state-space model framework for VSR that incorporates an innovative content-aware scanning mechanism.
Unlike rigid 1D sequential processing in conventional vision Mamba methods, our MambaVSR enables dynamic spatiotemporal interactions through the Shared Compass Construction (SCC) and the Content-Aware Sequentialization (CAS). 
Specifically, the SCC module constructs intra-frame semantic connectivity graphs via efficient sparse attention and generates adaptive spatial scanning sequences through spectral clustering. Building upon SCC, the CAS module effectively aligns and aggregates non-local similar content across multiple frames by interleaving temporal features along the learned spatial order.
To bridge global dependencies with local details, the Global-Local State Space Block (GLSSB) synergistically integrates window self-attention operations with SSM-based feature propagation, enabling high-frequency detail recovery under global dependency guidance.
Extensive experiments validate MambaVSR's superiority, outperforming the Transformer-based method by 0.58 dB PSNR on the REDS dataset with 55\% fewer parameters.
\end{abstract}

%% file: sec/1_intro.tex
\section{Introduction}
\label{sec:intro}
Video super-resolution (VSR) aims to reconstruct high-resolution (HR) video sequences from low-resolution (LR) inputs, with broad applications in live streaming~\cite{lu2018you}, video surveillance~\cite{zhang2010super}, and old film restoration~\cite{wan2022bringing}.
Different from the single image super-resolution (SISR), which primarily relies on the self-similarity within a single frame, VSR faces the core challenge of fully exploiting the highly correlated inter-frame dependencies across reference frames in a high-quality and efficient manner.

Recently, various deep learning-based VSR methods can be roughly categorized into two kinds depending on how they exploit temporal information: methods based on CNN-based methods~\cite{wang2019edvr,tian2020tdan,yi2019progressive-PFNL,chan2021basicvsr,chan2022basicvsr++,li2023MFPI} and Transformer-based methods~\cite{liang2024vrt,liang2022rvrt,tang2023ctvsr,TTVSR,shi2022rethinking,zhou2024miavsr}.
As shown in Figure~\ref{fig:intro}, CNN-based VSR methods typically rely on optical flow to mitigate inherent locality biases, but they often struggle with large displacements and complex motion in low-fps and low-quality videos, indicating persistent challenges in modeling non-local dependencies~\cite{ranjan2017optical-spynet}.
Although Transformers demonstrate superior non-local feature modeling through parallel self-attention on adjacent frame features, their deployment in long video scenarios is still limited by the quadratic computational complexity imposed by both spatial and temporal dimensions~\cite{liang2024vrt}.

Given a novel selective State Space Model (SSM) named Mamba~\cite{gu2023mamba,dao2024transformers-mamba2}, which uniquely combines linear computational complexity and a global receptive field, it is natural to replace traditional CNN and Transformer blocks with Mamba in VSR frameworks. However, as illustrated in Figure~\ref{fig:intro}, Mamba’s reliance on cumulative historical context during feature reconstruction enforces a rigid scanning strategy that disperses semantically similar features across distant sequence positions. This rigidity hinders effective inter-frame dependency capture despite its theoretically global receptive field. Moreover, the fixed scanning mechanism disrupts local high-frequency details~\cite{guo2024mambair,weng2025mamballie}, ultimately degrading VSR performance.

Unlike direct SSM adoption with fixed scanning paradigms, we propose MambaVSR, the first VSR framework to effectively leverage selective SSM through content-aware scanning mechanisms. 
Specifically, MambaVSR begins with the Shared Compass Construction (SCC), which employs an efficient sparse attention mechanism to establish intra-frame semantic graphs that capture salient feature dependencies through learned affinity relations, and applies spectral clustering to generate dynamic scanning order as a multi-stage shared compass, providing scanning guidance and enhancing efficiency.
Building upon the shared compass, the patch alignment operation and temporal feature interleaving along the learned order are employed to derive Content-Aware Sequentialization (CAS), which unifies globally similar features across misaligned frames to capture spatiotemporal long-range dependencies with low complexity.
Additionally, MambaVSR retains the feature window structure while adaptively concatenating window self-attention blocks of the same size, promoting the recovery of local high-frequency details under the guidance of global dependencies, with both components synergistically combined in the Global-Local State Space Block (GLSSB) to enhance reconstruction quality. 
Our approach explicitly addresses the challenges of inter-frame modeling in VSR. To the best of our knowledge, MambaVSR is the first framework to successfully integrate Mamba for this task.

Extensive experiments on benchmark datasets demonstrate the superior performance and efficiency of MambaVSR compared to state-of-the-art VSR approaches. In particular, MambaVSR outperforms the Transformer model VRT~\cite{liang2024vrt} by 0.58 dB in PSNR on the REDS dataset while using nearly 55\% fewer parameters, and it achieves a 0.22 dB improvement over the Swin-Transformer model PSRT-recurrent~\cite{shi2022rethinking} on the Vid4 dataset.

Our principal contributions can be summarized as follows:
\begin{itemize}[nosep]
  \item We propose MambaVSR, the first VSR framework based on selective SSM, leveraging content-aware scanning to achieve effective inter-frame feature modeling with linear complexity and global receptive field. 
  \item We propose SCC and CAS to capture long-range spatiotemporal dependencies, and introduce GLSSB to enhance the reconstruction of local high-frequency details guided by global contextual information.
  \item We conduct rigorous theoretical analysis and extensive experiments to validate the effectiveness of our adapted SSM for high-quality VSR, achieving superior qualitative and quantitative performance compared to state-of-the-art CNN-based and Transformer-based methods.
\end{itemize}

%% file: sec/2_RelatedWork.tex
\section{Related Work}
\label{sec:relatedwork}
\subsection{Video Super Resolution}
Existing VSR models can be roughly divided into two categories: CNN-based and Transformer-based methods. 
CNN-based methods generally rely on optical flow estimation to align adjacent frames, thereby extracting inter-frame information to recover missing details.
Representatively, BasicVSR and IconVSR~\cite{chan2021basicvsr} introduce bidirectional temporal propagation with optical flow–guided alignment.
BasicVSR++~\cite{chan2022basicvsr++} further enhances long-term motion modeling via second-order grid propagation and dynamic modulation.
However, due to the intrinsic locality bias of CNNs and the potential inaccuracies in flow estimation, these methods still struggle to capture non-local similarities effectively.

Rather than relying solely on insecure motion estimation to aggregate distant pixel information, Transformer-based methods achieve long-term modeling through self-attention mechanisms. 
Typically, VSR-Transformer~\cite{VSR-Transformer} employs spatiotemporal attention alongside bidirectional optical flow–based feedforward alignment. 
VRT~\cite{liang2024vrt} introduces temporal mutual attention for joint motion estimation and alignment. 
To expand the temporal receptive field limited by the quadratic complexity of global attention, PSRT~\cite{shi2022rethinking} combines bidirectional propagation with windowed attention. 
However, it faces an inherent trade-off between efficiency and receptive field: a large window incurs prohibitive computational costs, while restricted local attention fails to capture motion dependencies across windows.
\subsection{Vision Mamba}
In the field of Natural Language Processing, SSMs~\cite{gu2021combining} can efficiently capture long-range dependencies in language via state-space transformations. S4~\cite{gu2021efficiently} introduces a structured SSM with linear complexity, and Mamba~\cite{gu2023mamba} further enhances the performance through hardware design and selective parallel scanning (S6), and surpasses the Transformers~\cite{vaswani2017attention,viT} on long sequences.
However, Mamba is designed for 1D data and cannot be directly applied to visual data, which requires both global spatial context and local relational cues. 
Recently, Mamba has been extended in some low-level vision tasks~\cite{liu2024vmamba,guo2024mambair,guo2024mambairv2,weng2025mamballie,wu2024rainmamba,zhang2025vfimamba}. 
For example, VMamba~\cite{liu2024vmamba} utilizes the simultaneous multi-direction scanning to capture local textures and global structures.
VFIMamba~\cite{zhang2025vfimamba} rearranges the tokens from adjacent frames for S6.
MambaIRv2~\cite{guo2024mambairv2} proposes an attentive state-space equation for single-scan image unfolding.
RainMamba~\cite{wu2024rainmamba} introduces a Hilbert scanning mechanism to capture sequence-level local information. 

Nevertheless, how to deal with the high complexity of multi-directional scanning for video and the difficulty in modeling global similar inter-frame dependencies is still a significant challenge for Mamba.
Therefore, in this paper, we introduce a content-aware scanning SSM to improve the long-term modeling capabilities for VSR tasks while keeping the computational cost within an acceptable range.

%% file: sec/3_method.tex
\section{Method}
\subsection{Architecture}
Given $T$ low-resolution frames, the reconstruction pipeline comprises shallow feature extraction, cascaded bidirectional refinement, and feature reconstruction. Built upon BasicVSR++’s propagation backbone, MambaVSR integrates Global-Local State Space Groups, which include Shared Compass Construction and Content-Aware Sequentialization modules.

\subsection{Shared Compass Construction}
To reduce computational cost, the input frames are first downsampled via convolutional embedding layers. 
A dual-branch attention mechanism then computes a similarity matrix, capturing both dense and sparse dependencies. 
Graph-based spectral sorting is performed by constructing a Laplacian matrix and extracting the Fiedler vector, which yields a spatial scan order $\mathcal{O}$ through sorting. 
This order serves as shared prior guidance across refinement stages.

\subsection{Content-Aware Sequentialization}
Instead of recomputing spatial orders per frame, MambaVSR uses PatchAlignment to coarsely align multi-frame features, enabling reuse of the current frame’s order $\mathcal{O}$. 
The final scanning sequence ${\mathcal{S}}$ is produced by interleaving spatial features across aligned frames, ensuring temporal coherence and preserving local-global continuity.

\subsection{Global-Local State Space Block}
To balance long-range modeling and local detail preservation, pixel-level sequentialization is conducted within fixed spatial windows. 
Each window integrates self-attention for local structure recovery and state-space modeling for capturing global dependencies. Adaptive residual fusion is used to combine the two pathways effectively.

%% file: sec/4_experiments.tex
\section{Experiment}
\subsection{Experimental Setup}
\begin{table*}[!htp]
	\renewcommand{\arraystretch}{0.87} 
	\renewcommand{\tabcolsep}{8pt} 
    \begin{adjustbox}{center}
    \centering
    \begin{tabular}{c|c|c|c||cc|cc|cc}
    \toprule
    \multirow{2}{*}{Methods} & Frames &  Params & FLOPs & \multicolumn{2}{c}{REDS4} & \multicolumn{2}{c}{Vimeo-90K-T} & \multicolumn{2}{c}{Vid4}  \\
    & REDS/Vimeo & (M) & (T) & PSNR & SSIM & PSNR & SSIM & PSNR & SSIM    \\
    \midrule
    TOFlow~\cite{xue2019video} & 5/7 &-  &- & 27.98 & 0.7990 &33.08 &0.9054 & 25.89 & 0.7651 \\
    EDVR~\cite{wang2019edvr} & 5/- &20.6  &2.95 & 31.09 & 0.8800 & 36.09 & 0.9379 & 26.84& 0.8130\\
    PFNL~\cite{yi2019progressive-PFNL} &7/7 &3.0 &- &29.63 &0.8502 &36.14 &0.9363 & 26.73 & 0.8029 \\
    MuCAN~\cite{li2020mucan} & 5/- &-  &1.07 & 30.88 & 0.8750& - & - & -&- \\
    VSR-T~\cite{VSR-Transformer} & 5/- &32.6  &- &31.19 & 0.8815&  - & - & -&-\\
    PSRT-sliding~\cite{shi2022rethinking} & 5/- &14.8  &1.66 &  31.32 & 0.8834 & - & -& -& - \\
    PSRT-recurrent~\cite{shi2022rethinking} & 6/- &13.4  &2.43 & {31.85} & {0.8964}  &- & -& -& -\\
    VRT~\cite{liang2024vrt} & 6/- &30.7  &1.37 &31.59 & 0.8886  &- & -& -& - \\
    MambaVSR(Ours) & 6/- & 14.1  &2.46 & \textbf{31.95} & \textbf{0.8991}  &- & -& -& -  \\
    \midrule
    
    BasicVSR~\cite{chan2021basicvsr}     & 15/14 &6.3  &0.33 & 31.42 & 0.8909& 37.18 & 0.9450 & 27.24 & 0.8251\\
    IconVSR~\cite{chan2021basicvsr}      & 15/14  & 8.7 &0.51 & 31.67 & 0.8948& 37.47 & 0.9476 & 27.39 & 0.8279\\
    TTVSR\cite{TTVSR}&{50/-} &6.8  &0.61 &32.12&0.9021&-&-&-&-\\
    BasicVSR++~\cite{chan2022basicvsr++} & {30/14} &7.3  &0.39 & 32.39 & 0.9069& 37.79 & 0.9500 & 27.79 & 0.8400 \\
    RVRT~\cite{liang2022rvrt}       & {16/14} &10.8  &2.21 & {32.05} & {0.8974}& 38.15 & 0.9527 & 27.99 & 0.8462 \\
    CTVSR\cite{tang2023ctvsr}       & {16/14} &34.5  &0.63 & {32.25} & {0.9047}& - & - & 28.03 & 0.8487 \\
    PSRT-recurrent~\cite{shi2022rethinking} & 16/14 &13.4  &2.43 & 32.72 & 0.9106 &{38.27} & {0.9536}&{28.07}& {0.8485}\\
    VRT~\cite{liang2024vrt}              & 16/14  &30.7  &1.37 & 32.17 & 0.9002& 38.20 & 0.9530 & {27.93} & 0.8425\\
    MambaVSR(Ours)& 16/14 &14.1  &2.46 & \textbf{32.75} & \textbf{0.9110} & \textbf{38.33} & \textbf{0.9539} & \textbf{28.20} & \textbf{0.8514}\\
    \midrule
    IART~\cite{IART}              & 6/-  &13.4  &2.51 & 32.15 & 0.9110 &- & -&-&- \\
    MIA-VSR~\cite{zhou2024miavsr} & 6/- &16.5  & 1.61 & 32.01 & 0.8997 &- & -&-&- \\ 
    VideoGigaGAN~\cite{xu2025videogigagan}              & 10/-  &369  &1.37 & 30.46 & 0.8718 &- & -&-&- \\
    MambaVSR+(Ours)& 6/- &14.1  &2.54 & \textbf{32.17} & \textbf{0.9033}  &- & -&-&- \\
    \bottomrule
    \end{tabular}
  \end{adjustbox}
\caption{Quantitative comparison results (average PSNR/SSIM) on the REDS4~\cite{nah2019ntire}, Vimeo90K-T~\cite{xue2019video}, and Vid4~\cite{liu2013Vid4} datasets for the 4× video super-resolution task. All evaluations are conducted using NVIDIA GeForce RTX3090 GPUs, with FLOPs measured on a single 180×320 low-resolution frame. ``+'' denotes training with the advanced resampling module~\cite{IART}.}
\label{tab:sota}
\end{table*}
\noindent\textbf{Datasets and Training Details}. 
Following previous VSR studies~\cite{chan2022basicvsr++,shi2022rethinking}, we train our MambaVSR for 4× super-resolution under Bicubic (BI) degradation on the REDS~\cite{nah2019ntire} and Vimeo-90K~\cite{xue2019video} datasets, with Peak Signal-to-Noise Ratio (PSNR) and Structural Similarity Index Measure (SSIM)\cite{yang2020learning} between the ground truth and reconstructed videos serving as our quantitative metrics. 
Recent works~\cite{liang2024vrt,shi2022rethinking} have demonstrated that longer training sequences substantially enhance VSR performance. To ensure a fair comparison, we adopt the protocol of~\cite{shi2022rethinking}, training all VSR models on both 6-frame and 16-frame sequences from REDS and 14-frame sequences from Vimeo-90K. 
Specifically, models trained on 6-frame sequences undergo 300,000 iterations, whereas those on 16-frame sequences run for 600,000 iterations. Finally, the trained models are evaluated on three benchmark splits: REDS4~\cite{nah2019ntire}, Vimeo90K-T~\cite{xue2019video}, and Vid4~\cite{liu2013Vid4}.

\noindent\textbf{Parameter Settings.}
During training, we employ the Adam optimizer~\cite{kingma2014adam} with momentum parameters $\beta_{1}=0.9$ and $\beta_{2}=0.99$, and a Cosine Annealing learning‐rate scheduler~\cite{loshchilov2016sgdr}. 
The SpyNet~\cite{ranjan2017optical-spynet} utilizes pretrained weights and remains frozen for the first 5,000 iterations with a base learning rate of $2.5\times10^{-5}$, while other network parameters are optimized at an initial rate of $2\times10^{-4}$.
We set the batch size to 8 and use input patches of size $64\times64$.
To mitigate overfitting, we employ the widely used VSR data augmentation transformations of random horizontal flips, vertical flips, and $90^\circ$ rotations. 
The training loss adopts the Charbonnier penalty loss~\cite{lai2017deep}, which measures the reconstruction error between the enhanced frame $x_{SR}$ and its corresponding ground truth $x_{HR}$, and is formulated as:
\begin{equation}
    \label{loss}
    \mathcal{L}_{sr}=\sqrt{\|x_{HR}-x_{SR}\|^{2}+\varepsilon^2},
\end{equation}
where $\varepsilon$ is set to $10^{-3}$.
\subsection{Comparisons with State-of-the-Art Methods}
In the subsection, we compare MambaVSR with state‑of‑the‑art VSR methods: representative CNN‑based methods TOFlow~\cite{xue2019video}, EDVR~\cite{wang2019edvr}, MuCAN~\cite{li2020mucan}, BasicVSR~\cite{chan2021basicvsr}, IconVSR~\cite{chan2021basicvsr}, BasicVSR++~\cite{chan2022basicvsr++} and representative Transformer‑based methods VSR‑T~\cite{VSR-Transformer}, TTVSR~\cite{TTVSR}, RVRT~\cite{liang2022rvrt}, CTVSR~\cite{tang2023ctvsr}, PSRT~\cite{shi2022rethinking}, VRT~\cite{liang2024vrt}.



\begin{figure*}
\centering
\includegraphics[width=1.0\textwidth]{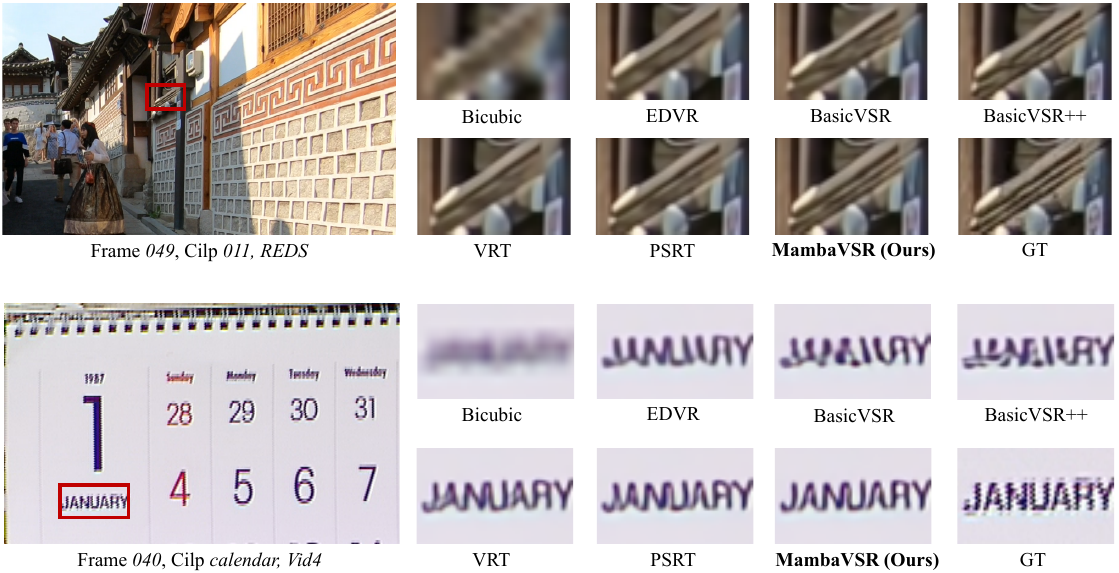}
\caption{Visual comparison results for $4\times$ VSR on REDS and Vid4 dataset. MambaVSR effectively restores realistic details.}
\label{fig:sotavisual}
\end{figure*}

The quantitative comparison results are as shown in Table~\ref{tab:sota}. MambaVSR demonstrates superior performance over both CNN and Transformer architectures across three benchmark datasets. 
Notably, our model achieves a 0.86 dB PSNR improvement over the CNN-based EDVR~\cite{wang2019edvr} on REDS4 under 6-frame training while reducing parameters by 32\%, and maintains a 0.58 dB advantage against the Transformer-based VRT~\cite{liang2024vrt} with 55\% fewer parameters under 16-frame training, validating that the proposed content-aware scanning strategy better balances global dependency modeling and efficiency.  
Furthermore, compared to the Swin‑Transformer‑based PSRT~\cite{shi2022rethinking}, MambaVSR obtains higher performance with comparable parameter count and FLOPs, evidencing its superior ability to exploit non-local spatiotemporal dependencies over windowed self‑attention.
Under a fair comparison utilizing the same advanced resampling module, MambaVSR also outperforms IART~\cite{IART}, achieving improvements of 0.02 dB in PSNR and 0.0023 in SSIM.
Visual quality comparisons are further provided in Figures~\ref{fig:sotavisual} to demonstrate MambaVSR's superior effectiveness against competing approaches. Specifically, where alternative methods generate blurred outputs with noticeable artifacts, MambaVSR maintains precise wall detailing with sharp edges and reproduces textual elements with high fidelity, consistently aligning with ground-truth references under diverse motion patterns and degradation conditions.
\begin{figure}[!htp]
\centering
\includegraphics[width=0.48\textwidth]{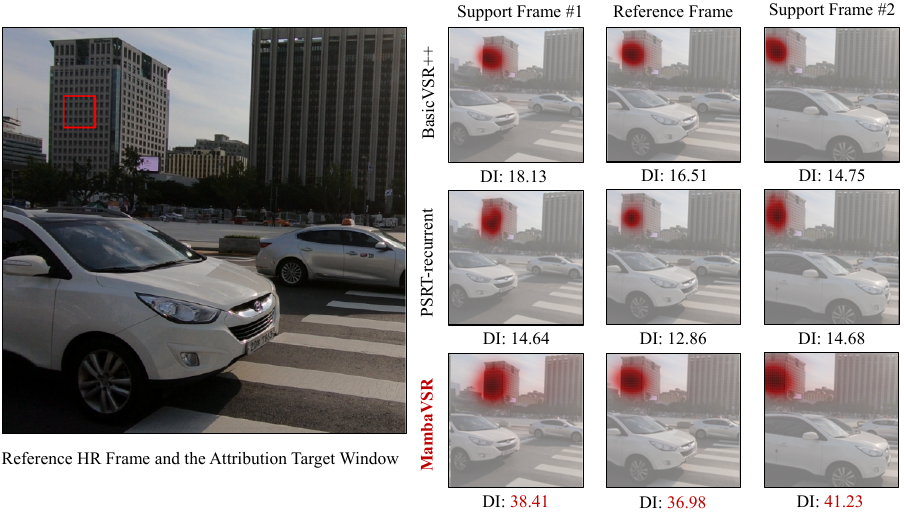}
\caption{Local Attribution Map (LAM)~\cite{gu2021interpreting} results and Diffusion Index (DI) for different networks. LAM visualizes the pixel-wise importance and receptive field size in the input LR frames for reconstructing the red-box region of the SR target frame. DI quantifies the spatial extent of relevant and utilized pixels.}
\label{fig:LAM}
\end{figure}

\begin{figure}[!h]
\centering
\includegraphics[width=0.45\textwidth]{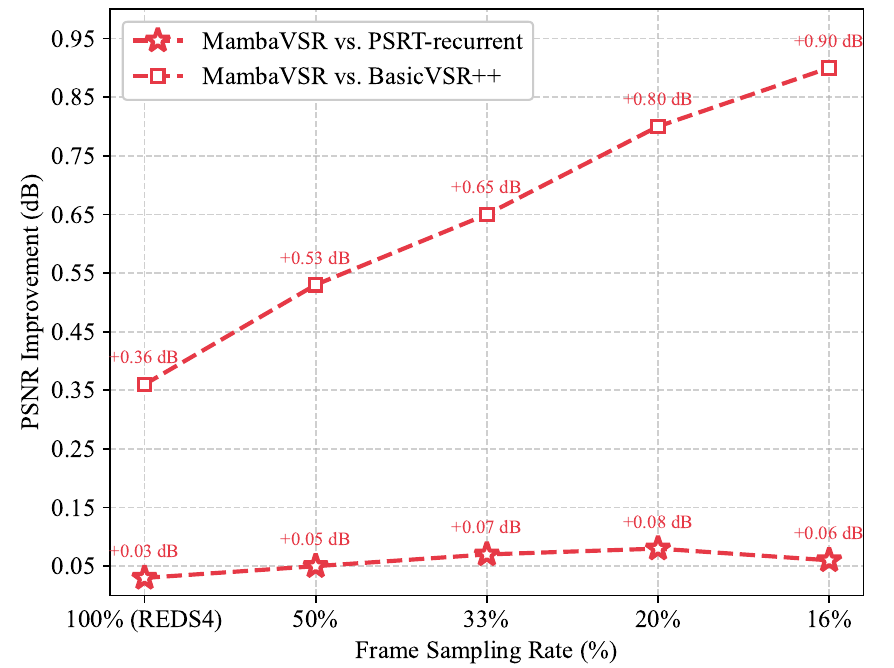}
\caption{PSNR Improvement of MambaVSR over BasicVSR++ and PSRT at decreasing temporal sampling rates.}
\label{fig:psnr_improve}
\end{figure}

To verify the superior long-range modeling capability of MambaVSR over other VSR models, we compare it with two representative linear complexity models: the CNN-based BasicVSR++ and the Swin Transformer-based PSRT.
As shown in Figure~\ref{fig:LAM}, MambaVSR activates more relevant pixels and achieves a higher DI than both PSRT and BasicVSR++, demonstrating its superior ability to aggregate relevant information from input frames by leveraging long-range spatiotemporal dependencies, thereby attaining a larger theoretical receptive field~\cite{chen2023activating}.
To further validate the receptive field advantage, we conduct progressive evaluations using large-motion test sets generated through temporal downsampling of REDS4~\cite{nah2019ntire}. As shown in Figure~\ref{fig:psnr_improve}, MambaVSR demonstrates greater PSNR improvements with increasing motion magnitude, validating the LAM analysis and further confirming its strong ability to exploit long-range spatiotemporal dependencies for better reconstruction performance.
\subsection{Ablation Studies}
In this subsection, we conduct a comprehensive evaluation of individual components in our proposed MambaVSR and perform comparisons with all models. 
For fair comparison, all models are trained under parameter configurations identical to our MambaVSR, using 6-frame video sequences from the REDS dataset~\cite{nah2019ntire} for 200,000 iterations. 
We employ two widely adopted benchmark test sets, REDS4 and Vid4, with quantitative evaluations conducted on them.
Specifically, we compare the proposed method in two ways to show its effectiveness for VSR.
First, we progressively remove the sequentialization modules in MambaVSR to evaluate how content-aware scanning impacts performance.
Subsequently, we selectively remove and add components within the GLSSB architecture, thereby demonstrating the superiority of the proposed GLSSB that leverages global information to guide local detail reconstruction.
\begin{table}
\centering
\begin{threeparttable}
\begin{adjustbox}{center}  
    \renewcommand{\arraystretch}{0.85} 
    \centering
    \scalebox{0.80}{\begin{tabular}{c|c||cc||cc}
    \toprule
    \multirow{2}{*}{Models} & Params &\multicolumn{2}{c}{REDS4} &\multicolumn{2}{c}{Vid4} \\
    &(M)& PSNR & SSIM  & PSNR & SSIM   \\
    \midrule
    Raster-based scanning& 13.91 & 31.68 & 0.8942 & 27.30 & 0.8299 \\
    Fielder-based scanning& 14.13 & 31.73 & 0.8945 & 27.32 & 0.8302 \\
    Content-Aware scanning& 14.13 & \textbf{31.82} & \textbf{0.8968} & \textbf{27.54} & \textbf{0.8392} \\
    \bottomrule
  \end{tabular}}
  \end{adjustbox}
  \end{threeparttable}
   \caption{Ablation studies on the proposed Content-Aware Sequentialization. } 
\label{tab:abCAS}
\end{table}

\begin{figure}
\centering
\includegraphics[width=0.48\textwidth]{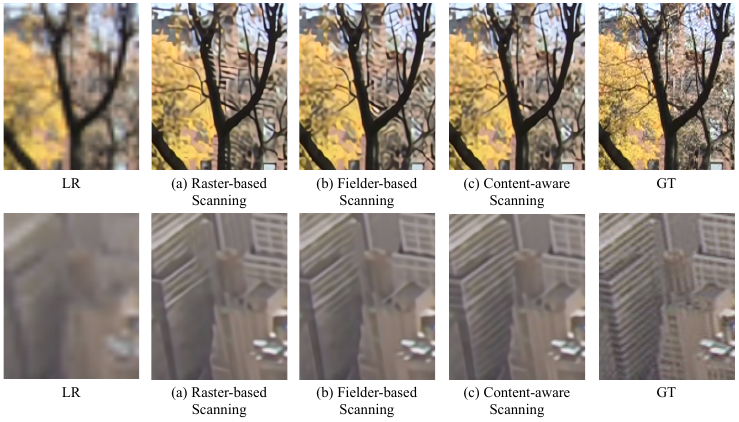}
\caption{Effectiveness of our proposed Content-Aware Scanning.}
\label{fig:casEffective}
\end{figure}

\begin{table}
\centering
\begin{threeparttable}
\begin{adjustbox}{center}  
    \centering
    \scalebox{0.80}{\begin{tabular}{c|c|c||cc}
    \toprule
    \multirow{2}{*}{Models} & Params & FLOPs &\multicolumn{2}{c}{REDS4} \\
    &(M)& (T)&PSNR & SSIM\\
    \midrule
    WFSAB& 8.93 &1.41 &31.41 & 0.8914  \\ 
    WFSAB-WFSAB & 13.4 &2.43 & {31.74} & {0.8955} \\
    WFSAB-GLSSM w/o $\gamma$& 14.12 & 2.46& {31.81} & {0.8966}\\
    WFSAB-GLSSM w/ $\gamma$& 14.13 & 2.46 &\textbf{31.82} & \textbf{0.8968}\\ 
    GLSSM-GLSSM & 14.73 & 2.50 & 31.63 & 0.8948 \\
    \bottomrule
  \end{tabular}}
  \end{adjustbox}
  \end{threeparttable}
 \caption{Ablation studies on the proposed Global-Local State Space Block. } 
\label{tab:abGLSSB}
\end{table}

\noindent\textbf{Effectiveness of CAS.}
For quantitative comparisons, as shown in Table~\ref{tab:abCAS}, we use the MambaVSR model with fixed raster scanning as the baseline. We then gradually incorporate a fielder-based scanning algorithm with single-frame content-aware capability, achieving a 0.05 dB PSNR improvement on the REDS4 dataset with only an additional parameter of 0.2 M. To further enhance temporal content awareness, we introduce a temporal interleaving mechanism, ultimately developing our Content-aware scanning method, which yields a 0.24 dB PSNR gain over the baseline on the Vid4 dataset.

For qualitative comparisons, as shown in Figure~\ref{fig:casEffective}, fixed raster scanning introduces aliasing artifacts along tree branches and blurred floor textures. While field-based scanning moderately reduces these artifacts through per-frame content awareness, our content-aware scanning method effectively eliminates them by dynamically aggregating multi-frame information.
Together, the above analyses demonstrate the superiority of our content-aware scanning approach in adaptively leveraging inter-frame information based on video content.

\noindent\textbf{Effectiveness of GLSSB.}
Our GLSSB leverages long-range dependencies to guide local high-frequency detail restoration. To validate the design, we decompose GLSSB into WFSAB and GLSSM modules and investigate the impact of their connection via a learnable parameter $\gamma$. 
As shown in Table~\ref{tab:abGLSSB}, the mixed WFSAB-GLSSM structure without $\gamma$ outperforms the local-focused WFSAB-WFSAB and global-focused GLSSM-GLSSM structures by 0.07 dB and 0.18 dB PSNR on REDS4, respectively. The introduction of $\gamma$ further boosts performance by 0.01 dB, collectively demonstrating GLSSB's effectiveness.

%% file: sec/5_conclusion.tex
\section{Conclusion}
In this paper, We present the first SSM framework for VSR named MambaVSR, which addresses the challenge of modeling non-local dependencies across misaligned frames with high efficiency.
MambaVSR introduces a content-aware scanning paradigm that enables dynamic spatiotemporal interactions beyond the rigid 1D sequential processing of conventional vision Mamba models. 
Specifically, the Shared Compass Construction (SCC) module builds intra-frame semantic graphs and generates adaptive spatial scanning sequences, while the Content-Aware Sequentialization (CAS) module aligns and aggregates non-local content across frames.
The Global-Local State Space Block (GLSSB) further integrates window self-attention with SSM-based feature propagation to recover fine details under global guidance. 
Both quantitative and qualitative results show that MambaVSR outperforms state-of-the-art methods on several benchmark datasets


%% file: main.bbl
\begin{thebibliography}{41}
\providecommand{\natexlab}[1]{#1}
\providecommand{\url}[1]{\texttt{#1}}
\expandafter\ifx\csname urlstyle\endcsname\relax
  \providecommand{\doi}[1]{doi: #1}\else
  \providecommand{\doi}{doi: \begingroup \urlstyle{rm}\Url}\fi

\bibitem[Cao et~al.(2021)Cao, Li, Zhang, and Van~Gool]{VSR-Transformer}
Jiezhang Cao, Yawei Li, Kai Zhang, and Luc Van~Gool.
\newblock Video super-resolution transformer.
\newblock \emph{arXiv preprint arXiv:2106.06847}, 2021.

\bibitem[Chan et~al.(2021)Chan, Wang, Yu, Dong, and Loy]{chan2021basicvsr}
Kelvin~CK Chan, Xintao Wang, Ke Yu, Chao Dong, and Chen~Change Loy.
\newblock {BasicVSR}: The search for essential components in video super-resolution and beyond.
\newblock In \emph{Proceedings of the IEEE/CVF Conference on Computer Vision and Pattern Recognition}, pages 4947--4956, 2021.

\bibitem[Chan et~al.(2022)Chan, Zhou, Xu, and Loy]{chan2022basicvsr++}
Kelvin~CK Chan, Shangchen Zhou, Xiangyu Xu, and Chen~Change Loy.
\newblock {BasicVSR}++: Improving video super-resolution with enhanced propagation and alignment.
\newblock In \emph{Proceedings of the IEEE/CVF Conference on Computer Vision and Pattern Recognition}, pages 5972--5981, 2022.

\bibitem[Chen et~al.(2023)Chen, Wang, Zhou, Qiao, and Dong]{chen2023activating}
Xiangyu Chen, Xintao Wang, Jiantao Zhou, Yu Qiao, and Chao Dong.
\newblock Activating more pixels in image super-resolution transformer.
\newblock In \emph{Proceedings of the IEEE/CVF Conference on Computer Vision and Pattern Recognition}, pages 22367--22377, 2023.

\bibitem[Dao and Gu(2024)]{dao2024transformers-mamba2}
Tri Dao and Albert Gu.
\newblock Transformers are ssms: Generalized models and efficient algorithms through structured state space duality.
\newblock \emph{arXiv preprint arXiv:2405.21060}, 2024.

\bibitem[Dosovitskiy et~al.(2020)Dosovitskiy, Beyer, Kolesnikov, Weissenborn, Zhai, Unterthiner, Dehghani, Minderer, Heigold, Gelly, et~al.]{viT}
Alexey Dosovitskiy, Lucas Beyer, Alexander Kolesnikov, Dirk Weissenborn, Xiaohua Zhai, Thomas Unterthiner, Mostafa Dehghani, Matthias Minderer, Georg Heigold, Sylvain Gelly, et~al.
\newblock An image is worth 16x16 words: Transformers for image recognition at scale.
\newblock \emph{arXiv preprint arXiv:2010.11929}, 2020.

\bibitem[Gu and Dao(2023)]{gu2023mamba}
Albert Gu and Tri Dao.
\newblock Mamba: Linear-time sequence modeling with selective state spaces.
\newblock \emph{arXiv preprint arXiv:2312.00752}, 2023.

\bibitem[Gu et~al.(2021{\natexlab{a}})Gu, Goel, and R{\'e}]{gu2021efficiently}
Albert Gu, Karan Goel, and Christopher R{\'e}.
\newblock Efficiently modeling long sequences with structured state spaces.
\newblock \emph{arXiv preprint arXiv:2111.00396}, 2021{\natexlab{a}}.

\bibitem[Gu et~al.(2021{\natexlab{b}})Gu, Johnson, Goel, Saab, Dao, Rudra, and R{\'e}]{gu2021combining}
Albert Gu, Isys Johnson, Karan Goel, Khaled Saab, Tri Dao, Atri Rudra, and Christopher R{\'e}.
\newblock Combining recurrent, convolutional, and continuous-time models with linear state space layers.
\newblock In \emph{Advances in Neural Information Processing Systems}, pages 572--585, 2021{\natexlab{b}}.

\bibitem[Gu and Dong(2021)]{gu2021interpreting}
Jinjin Gu and Chao Dong.
\newblock Interpreting super-resolution networks with local attribution maps.
\newblock In \emph{Proceedings of the IEEE/CVF Conference on Computer Vision and Pattern Recognition}, pages 9199--9208, 2021.

\bibitem[Guo et~al.(2024{\natexlab{a}})Guo, Guo, Zha, Zhang, Li, Dai, Xia, and Li]{guo2024mambairv2}
Hang Guo, Yong Guo, Yaohua Zha, Yulun Zhang, Wenbo Li, Tao Dai, Shu-Tao Xia, and Yawei Li.
\newblock {MambaIRv2}: Attentive state space restoration.
\newblock \emph{arXiv preprint arXiv:2411.15269}, 2024{\natexlab{a}}.

\bibitem[Guo et~al.(2024{\natexlab{b}})Guo, Li, Dai, Ouyang, Ren, and Xia]{guo2024mambair}
Hang Guo, Jinmin Li, Tao Dai, Zhihao Ouyang, Xudong Ren, and Shu-Tao Xia.
\newblock {MambaIR}: A simple baseline for image restoration with state-space model.
\newblock In \emph{Proceedings of the European Conference on Computer Vision}, pages 222--241, 2024{\natexlab{b}}.

\bibitem[Kingma and Ba(2014)]{kingma2014adam}
Diederik~P Kingma and Jimmy Ba.
\newblock {Adam}: A method for stochastic optimization.
\newblock \emph{arXiv preprint arXiv:1412.6980}, 2014.

\bibitem[Lai et~al.(2017)Lai, Huang, Ahuja, and Yang]{lai2017deep}
WeiSheng Lai, JiaBin Huang, Narendra Ahuja, and MingHsuan Yang.
\newblock Deep laplacian pyramid networks for fast and accurate super-resolution.
\newblock In \emph{Proceedings of the IEEE Conference on Computer Vision and Pattern Recognition}, pages 624--632, 2017.

\bibitem[Li et~al.(2023)Li, Zhang, Liu, Lei, and Li]{li2023MFPI}
Fei Li, Linfeng Zhang, Zikun Liu, Juan Lei, and Zhenbo Li.
\newblock Multi-frequency representation enhancement with privilege information for video super-resolution.
\newblock In \emph{Proceedings of the IEEE/CVF International Conference on Computer Vision}, pages 12814--12825, 2023.

\bibitem[Li et~al.(2020)Li, Tao, Guo, Qi, Lu, and Jia]{li2020mucan}
Wenbo Li, Xin Tao, Taian Guo, Lu Qi, Jiangbo Lu, and Jiaya Jia.
\newblock {MuCAN}: Multi-correspondence aggregation network for video super-resolution.
\newblock In \emph{Proceedings of the European Conference on Computer Vision}, pages 335--351, 2020.

\bibitem[Liang et~al.(2022)Liang, Fan, Xiang, Ranjan, Ilg, Green, Cao, Zhang, Timofte, and Gool]{liang2022rvrt}
Jingyun Liang, Yuchen Fan, Xiaoyu Xiang, Rakesh Ranjan, Eddy Ilg, Simon Green, Jiezhang Cao, Kai Zhang, Radu Timofte, and Luc~V Gool.
\newblock Recurrent video restoration transformer with guided deformable attention.
\newblock In \emph{Advances in Neural Information Processing Systems}, pages 378--393, 2022.

\bibitem[Liang et~al.(2024)Liang, Cao, Fan, Zhang, Ranjan, Li, Timofte, and Van~Gool]{liang2024vrt}
Jingyun Liang, Jiezhang Cao, Yuchen Fan, Kai Zhang, Rakesh Ranjan, Yawei Li, Radu Timofte, and Luc Van~Gool.
\newblock {VRT}: A video restoration transformer.
\newblock \emph{IEEE Transactions on Image Processing}, 33:\penalty0 2171--2182, 2024.

\bibitem[Liu and Sun(2013)]{liu2013Vid4}
Ce Liu and Deqing Sun.
\newblock On bayesian adaptive video super resolution.
\newblock \emph{IEEE Transactions on Pattern Analysis and Machine Intelligence}, 36\penalty0 (2):\penalty0 346--360, 2013.

\bibitem[Liu et~al.(2022)Liu, Yang, Fu, and Qian]{TTVSR}
Chengxu Liu, Huan Yang, Jianlong Fu, and Xueming Qian.
\newblock Learning trajectory-aware transformer for video super-resolution.
\newblock In \emph{Proceedings of the IEEE/CVF Conference on Computer Vision and Pattern Recognition}, pages 5687--5696, 2022.

\bibitem[Liu et~al.(2024)Liu, Tian, Zhao, Yu, Xie, Wang, Ye, and Liu]{liu2024vmamba}
Yue Liu, Yunjie Tian, Yuzhong Zhao, Hongtian Yu, Lingxi Xie, Yaowei Wang, Qixiang Ye, and Yunfan Liu.
\newblock {VMamba}: Visual state space model.
\newblock \emph{arXiv preprint arXiv:2401.10166}, 2024.

\bibitem[Loshchilov and Hutter(2016)]{loshchilov2016sgdr}
Ilya Loshchilov and Frank Hutter.
\newblock {SGDR}: Stochastic gradient descent with warm restarts.
\newblock \emph{arXiv preprint arXiv:1608.03983}, 2016.

\bibitem[Lu et~al.(2018)Lu, Xia, Heo, and Wigdor]{lu2018you}
Zhicong Lu, Haijun Xia, Seongkook Heo, and Daniel Wigdor.
\newblock You watch, you give, and you engage: a study of live streaming practices in china.
\newblock In \emph{Proceedings of the CHI Conference on Human Factors in Computing Systems}, pages 1--13, 2018.

\bibitem[Nah et~al.(2019)Nah, Baik, Hong, Moon, Son, Timofte, and Mu~Lee]{nah2019ntire}
Seungjun Nah, Sungyong Baik, Seokil Hong, Gyeongsik Moon, Sanghyun Son, Radu Timofte, and Kyoung Mu~Lee.
\newblock {NTIRE} 2019 challenge on video deblurring and super-resolution: Dataset and study.
\newblock In \emph{Proceedings of the IEEE/CVF Conference on Computer Vision and Pattern Recognition Workshops}, 2019.

\bibitem[Ranjan and Black(2016)]{ranjan2017optical-spynet}
Anurag Ranjan and Michael~J. Black.
\newblock Optical flow estimation using a spatial pyramid network.
\newblock In \emph{Proceedings of the IEEE Conference on Computer Vision and Pattern Recognition}, pages 2720--2729, 2016.

\bibitem[Shi et~al.(2022)Shi, Gu, Xie, Wang, Yang, and Dong]{shi2022rethinking}
Shuwei Shi, Jinjin Gu, Liangbin Xie, Xintao Wang, Yujiu Yang, and Chao Dong.
\newblock Rethinking alignment in video super-resolution transformers.
\newblock In \emph{Advances in Neural Information Processing Systems}, pages 36081--36093, 2022.

\bibitem[Tang et~al.(2024)Tang, Lu, Liu, Li, Dai, and Ding]{tang2023ctvsr}
Jun Tang, Chenyan Lu, Zhengxue Liu, Jiale Li, Hang Dai, and Yong Ding.
\newblock {CTVSR}: Collaborative spatial–temporal transformer for video super-resolution.
\newblock \emph{IEEE Transactions on Circuits and Systems for Video Technology}, 34\penalty0 (6):\penalty0 5018--5032, 2024.

\bibitem[Tian et~al.(2020)Tian, Zhang, Fu, and Xu]{tian2020tdan}
Yapeng Tian, Yulun Zhang, Yun Fu, and Chenliang Xu.
\newblock {TDAN}: Temporally-deformable alignment network for video super-resolution.
\newblock In \emph{Proceedings of the IEEE/CVF Conference on Computer Vision and Pattern Recognition}, pages 3360--3369, 2020.

\bibitem[Vaswani et~al.(2017)Vaswani, Shazeer, Parmar, Uszkoreit, Jones, Gomez, Kaiser, and Polosukhin]{vaswani2017attention}
Ashish Vaswani, Noam Shazeer, Niki Parmar, Jakob Uszkoreit, Llion Jones, Aidan~N Gomez, {\L}ukasz Kaiser, and Illia Polosukhin.
\newblock Attention is all you need.
\newblock In \emph{Advances in Neural Information Processing Systems}, page 6000–6010, 2017.

\bibitem[Wan et~al.(2022)Wan, Zhang, Chen, and Liao]{wan2022bringing}
Ziyu Wan, Bo Zhang, Dongdong Chen, and Jing Liao.
\newblock Bringing old films back to life.
\newblock In \emph{Proceedings of the IEEE/CVF Conference on Computer Vision and Pattern Recognition}, pages 17694--17703, 2022.

\bibitem[Wang et~al.(2019)Wang, Chan, Yu, Dong, and Change~Loy]{wang2019edvr}
Xintao Wang, Kelvin~CK Chan, Ke Yu, Chao Dong, and Chen Change~Loy.
\newblock {EDVR}: Video restoration with enhanced deformable convolutional networks.
\newblock In \emph{Proceedings of the IEEE/CVF Conference on Computer Vision and Pattern Recognition Workshops}, pages 1954--1963, 2019.

\bibitem[Weng et~al.(2025)Weng, Yan, Tai, Qian, Yang, and Li]{weng2025mamballie}
Jiangwei Weng, Zhiqiang Yan, Ying Tai, Jianjun Qian, Jian Yang, and Jun Li.
\newblock {MambaLLIE}: Implicit retinex-aware low light enhancement with global-then-local state space.
\newblock In \emph{Advances in Neural Information Processing Systems}, pages 27440--27462, 2025.

\bibitem[Wu et~al.(2024)Wu, Yang, Xu, Wang, Zhou, and Zhu]{wu2024rainmamba}
Hongtao Wu, Yijun Yang, Huihui Xu, Weiming Wang, Jinni Zhou, and Lei Zhu.
\newblock {RainMamba}: Enhanced locality learning with state space models for video deraining.
\newblock In \emph{Proceedings of the ACM International Conference on Multimedia}, pages 7881--7890, 2024.

\bibitem[Xu et~al.(2024)Xu, Yu, Wang, Mi, and Yao]{IART}
Kai Xu, Ziwei Yu, Xin Wang, Michael~Bi Mi, and Angela Yao.
\newblock Enhancing video super-resolution via implicit resampling-based alignment.
\newblock In \emph{Proceedings of the IEEE/CVF Conference on Computer Vision and Pattern Recognition}, pages 2546--2555, 2024.

\bibitem[Xu et~al.(2025)Xu, Park, Zhang, Zhou, Shechtman, Liu, Huang, and Liu]{xu2025videogigagan}
Yiran Xu, Taesung Park, Richard Zhang, Yang Zhou, Eli Shechtman, Feng Liu, Jia-Bin Huang, and Difan Liu.
\newblock Videogigagan: Towards detail-rich video super-resolution.
\newblock In \emph{Proceedings of the Computer Vision and Pattern Recognition Conference}, pages 2139--2149, 2025.

\bibitem[Xue et~al.(2019)Xue, Chen, Wu, Wei, and Freeman]{xue2019video}
Tianfan Xue, Baian Chen, Jiajun Wu, Donglai Wei, and William~T Freeman.
\newblock Video enhancement with task-oriented flow.
\newblock \emph{International Journal of Computer Vision}, 127:\penalty0 1106--1125, 2019.

\bibitem[Yang et~al.(2020)Yang, Yang, Fu, Lu, and Guo]{yang2020learning}
Fuzhi Yang, Huan Yang, Jianlong Fu, Hongtao Lu, and Baining Guo.
\newblock Learning texture transformer network for image super-resolution.
\newblock In \emph{Proceedings of the IEEE/CVF Conference on Computer Vision and Pattern Recognition}, pages 5791--5800, 2020.

\bibitem[Yi et~al.(2019)Yi, Wang, Jiang, Jiang, and Ma]{yi2019progressive-PFNL}
Peng Yi, Zhongyuan Wang, Kui Jiang, Junjun Jiang, and Jiayi Ma.
\newblock Progressive fusion video super-resolution network via exploiting non-local spatio-temporal correlations.
\newblock In \emph{Proceedings of the IEEE/CVF International Conference on Computer Vision}, pages 3106--3115, 2019.

\bibitem[Zhang et~al.(2024)Zhang, Liu, Cui, Zhao, Ma, and Wang]{zhang2025vfimamba}
Guozhen Zhang, Chuxnu Liu, Yutao Cui, Xiaotong Zhao, Kai Ma, and Limin Wang.
\newblock {VFIMamba}: Video frame interpolation with state space models.
\newblock In \emph{Advances in Neural Information Processing Systems}, pages 107225--107248, 2024.

\bibitem[Zhang et~al.(2010)Zhang, Zhang, Shen, and Li]{zhang2010super}
Liangpei Zhang, Hongyan Zhang, Huanfeng Shen, and Pingxiang Li.
\newblock A super-resolution reconstruction algorithm for surveillance images.
\newblock \emph{Signal Processing}, 90\penalty0 (3):\penalty0 848--859, 2010.

\bibitem[Zhou et~al.(2024)Zhou, Zhang, Zhao, Wang, Li, and Gu]{zhou2024miavsr}
Xingyu Zhou, Leheng Zhang, Xiaorui Zhao, Keze Wang, Leida Li, and Shuhang Gu.
\newblock Video super-resolution transformer with masked inter\&intra-frame attention.
\newblock In \emph{Proceedings of the IEEE/CVF Conference on Computer Vision and Pattern Recognition}, pages 25399--25408, 2024.

\end{thebibliography}
